\documentclass[letterpaper]{article} 
\usepackage{aaai2026}  
\usepackage{times}  
\usepackage{helvet}  
\usepackage{courier}  
\usepackage[hyphens]{url}  
\usepackage{graphicx} 
\usepackage{natbib}  
\usepackage{caption} 
\usepackage{algorithm}
\usepackage{algorithmic}
\usepackage{amsmath}
\usepackage{booktabs}
\usepackage{multirow}
\usepackage{array}
\usepackage{amssymb}

\usepackage{newfloat}
\usepackage{listings}
\DeclareCaptionStyle{ruled}{labelfont=normalfont,labelsep=colon,strut=off} 
\floatstyle{ruled}
\newfloat{listing}{tb}{lst}{}
\floatname{listing}{Listing}
\title{CARD: A Cache-Assisted Parallel Speculative Decoding Framework via Query-and-Correct Paradigm for Accelerating LLM Inference }
\author {
    Enyu Zhou\textsuperscript{\rm 1},
    Kai Sheng\textsuperscript{\rm 1},
    Hao Chen\textsuperscript{\rm 2},
    Xin He\textsuperscript{\rm 1}
}
\affiliations {
    \textsuperscript{\rm 1} Guangzhou Institute of Technology, Xidian University, \\
    \textsuperscript{\rm 2}CSEE, Hunan University\\
       \{zhouenyu\}@stu.xidian.edu.cn\\
     \{kaisheng, hexin\}@xidian.edu.cn, haochen@hnu.edu.cn
}

\usepackage{bibentry}

\begin{document}

\maketitle

\begin{abstract}

Speculative decoding (SD), where a draft model provides multiple candidate tokens for the target model to verify in parallel, has demonstrated significant potential for accelerating LLM inference. Yet, existing SD approaches adhere to a strict ``draft-then-verify'' paradigm, enforcing a sequential process that hampers performance and constrains the draft model’s capacity. Moreover, rejecting a token in the candidate sequence invalidates all subsequent tokens, leading to wasted computation during drafting.

To overcome these limitations, we propose a cache-assisted parallel speculative decoding framework called CARD, which employs a novel \textit{``query-and-correct''} paradigm. Our approach decouples drafting from verification: the draft model populates a shared cache with candidate tokens, while the target model concurrently refines the draft's trajectory. This enables inference at near-draft-speed, effectively leveraging the draft model’s efficiency without additional fine-tuning. Experimental results show that CARD significantly outperforms existing state-of-the-art methods, achieving up to a 4.83× acceleration over vanilla autoregressive decoding, with no fine-tuning required for either models.
\end{abstract}

\begin{links}
    \link{Code}{https://github.com/hunzhizi/CARD}
\end{links}

\section{Introduction}

\begin{figure}[t]
 \includegraphics[width=\columnwidth]{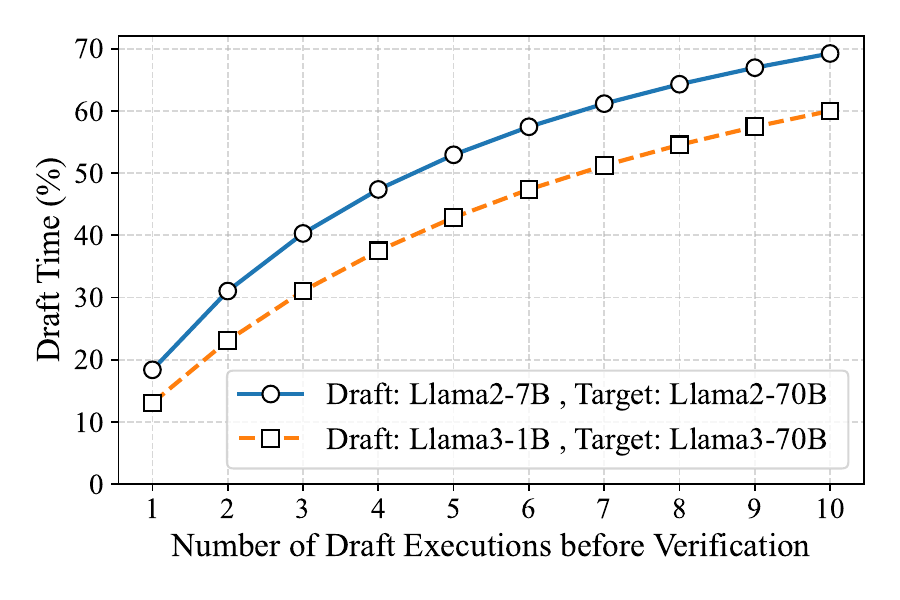}
 \caption{The ratio of draft model execution time to total inference time, highlighting the "mutual waiting" problem as the draft length $r$ increases.}
 \label{fig:introduction_motivation}
\end{figure}

Modern large language models (LLMs) such as ChatGPT-o1 and DeepSeek-R1 \citep{guo2025deepseek} demonstrate remarkable capabilities across diverse tasks. However, their widespread deployment is critically hindered by high latency inherent in autoregressive generation, where tokens are produced sequentially, resulting in a substantial performance bottleneck. This latency challenge significantly impacts various latency-critical deployment scenarios, including chatbots \citep{hariri2023unlocking}, search engines \citep{xiong2024search, caramancion2024large}, and personalized recommendation systems \citep{hariri2023unlocking, yang2023palr}.

To address this challenge, Speculative Decoding (SD) has emerged as a promising acceleration technique \citep{leviathan2023fast, chen2023accelerating, sun2023spectr}. SD operates under a ``draft-then-verify'' paradigm in which a smaller draft model generates candidate tokens in parallel, enabling a larger and more accurate target model to verify these candidates concurrently. This approach leverages the observation that large language models are predominantly memory-bound during inference, enabling lossless acceleration by harnessing additional computational resources.

However, existing SD frameworks are trapped in a fundamental dilemma, forcing a trade-off between drafting efficiency and the computational burden on the target model.
\begin{itemize}
\item  \textbf{On one hand, simple chain-based drafting is inefficient.} Methods such as PEARL \citep{liu2025pearl}, Lookahead \citep{Zhao2023LookaheadAI}, and SD \citep{leviathan2023fast} are easy to implement, but rejecting a single token invalidates the entire subsequent chain, leading to wasted computation.
\item  \textbf{On the other hand, complex tree-based drafting overloads the target model.} To improve the acceptance rate, advanced methods like EAGLE \citep{li2024eagle}, OPT-Tree \citep{wang2025opt}, and Ouroboros \citep{zhao2024ouroboros} generate sophisticated token trees. While this increases the likelihood of token acceptance, it significantly raises the verification workload on the resource-intensive target model, creating a new bottleneck.
\end{itemize}

The root of this dilemma lies in the strict sequential dependency between drafting and verification phases inherent in current SD designs. As shown in \ref{fig:introduction_motivation},this dependency results in a critical \textit{mutual waiting} problem: the powerful target model often remains idle, waiting for the draft model to finish, leading to severe underutilization of hardware. To minimize waiting time, state-of-the-art methods typically employ extremely lightweight draft models (e.g., a single MLP layer in Medusa \citep{cai2024medusa}), which degrade draft quality and limit overall speedup. Consequently, the core promise of SD is fundamentally constrained by this rigid sequential architecture.

To overcome this bottleneck, we propose CARD, a novel speculative decoding framework that rethinks the interaction paradigm by \textit{parallelizing the draft and target models}. CARD decouples these processes, eliminating the sequential dependency and enabling parallel execution. The key enabling idea is that CARD employs a novel \textit{'query-and-correct'} paradigm: the draft model continuously generates potential token sequences into a shared cache, while the target model queries this cache to verify and, if necessary, correct the most promising paths. This parallel design combines the strengths of both approaches: it achieves the high token acceptance rates characteristic of tree-based methods, without overburdening the target model, while maintaining the efficiency of chain-based verification. 

In summary, our main contributions are as follows.
\begin{itemize}
\item  \textbf{Parallel Execution Framework.} A cache-supported architecture that breaks the sequential dependency of traditional SD, enabling simultaneous execution of draft and target models and eliminating the mutual waiting problem.
\item  \textbf{Adaptive Caching Mechanism.} A dynamic cache that preserves historical generation states, allowing the framework to adaptively select optimal draft paths and maximize token acceptance.
\item  \textbf{Computation Offloading.} By shifting the complex exploration of token trees to the parallel draft model, CARD attains the high throughput of tree-based decoding while ensuring the target model only performs efficient chain-based verification.

\item \textbf{Extensive Evaluation.} We conduct comprehensive experiments demonstrating CARD's superiority over state-of-the-art baselines, achieving average performance gains of 1.30× over PEARL and 1.39× over Ouroboros. Notably, compared to EAGLE-3, CARD reduces the computational load on the target model by 6.86×. Additionally, CARD delivers up to 4.83× acceleration compared to vanilla autoregressive decoding, without requiring any fine-tuning of either the draft or target models. 
\end{itemize}

\begin{figure*}[t]
  \includegraphics[width=0.97\linewidth]{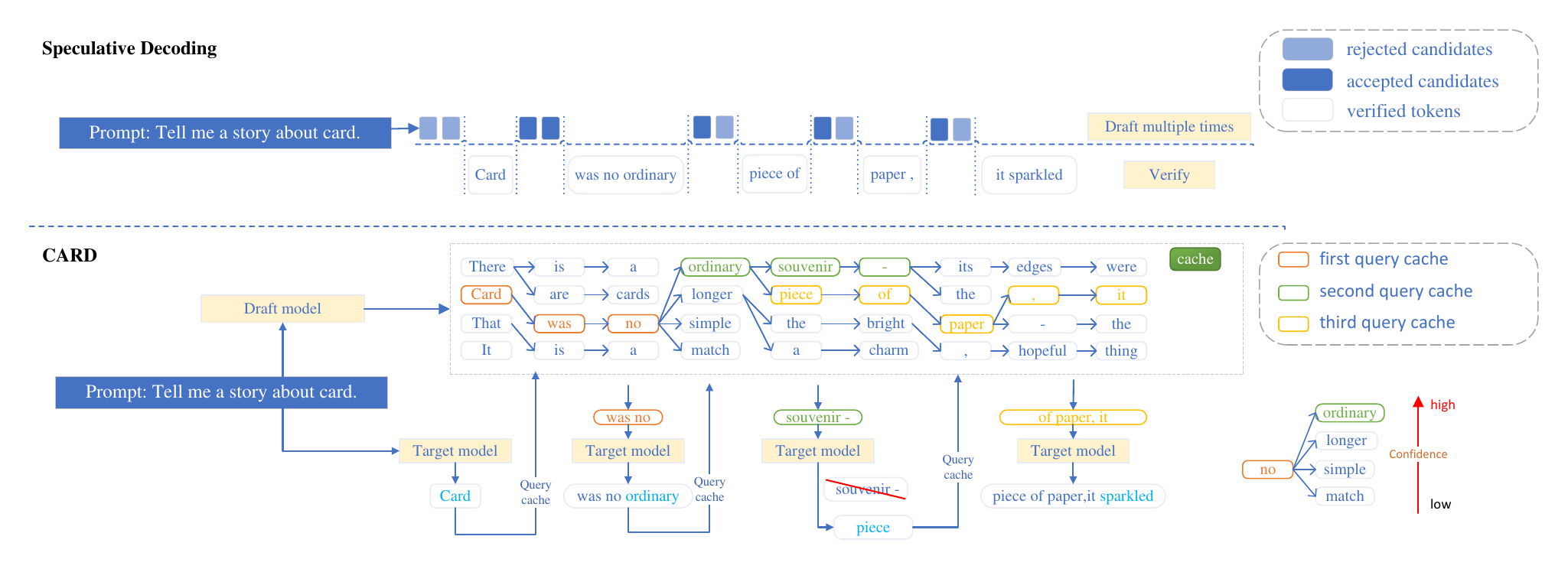}
  \caption{Overview of CARD.This diagram illustrates the three-step CARD inference process, assuming a 3:1 speed ratio between the draft and target models.Step 1: The target model queries the cache with the verified token “Card” and hits. It selects the highest-confidence path “was no,” which is accepted. The draft model prunes unrelated paths and continues from “Card,” while the target model samples the next token “ordinary.”Step 2: With “was no ordinary,” the target model again hits in the cache and selects “souvenir -,” which is rejected. The draft model prunes to “ordinary,” and the target model samples “piece.”Step 3: The target model queries “piece,” hits the cache, and selects “of paper, it,” which is accepted. The draft model prunes to “piece,” and the next token “sparkled” is sampled.}
  \label{fig:CARD_overview}
\end{figure*}

\section{CARD Design}

In CARD, we introduce a novel cache-assisted speculative decoding method  that breaks the traditional ``draft-then-verify'' paradigm, enabling the target model to perform continuous inference. As illustrated in Figure~\ref{fig:CARD_overview}, CARD replaces the conventional approach with a new \textit{query-and-correct''} paradigm. This approach allows the draft and target models to operate in parallel. Specifically, CARD accelerates the target model by caching its generation space and dynamically correcting the draft model’s generation direction after verification. This methodology serves as a generalizable framework that could potentially incorporate database retrieval or other caching strategies for the target model.

Within the speculative decoding domain, CARD employs a consistent strategy by using a draft model to implement caching for the target model. The process involves the draft model continuously generating candidate tokens to form potential outputs, from which a subset is selected as the cache for the target model. Before each inference step, the target model queries this cache and proceeds with inference only on the highest-weighted candidate token sequence. After verifying the generated tokens, CARD corrects the draft model’s generation direction accordingly, enabling it to produce high-quality tokens in a continuous manner. \textit{Notably, despite adopting this new paradigm, CARD preserves the sampling mechanisms of conventional speculative decoding, thereby ensuring lossless acceleration.}

\subsection{Preliminary}

\textbf{Generation space} $\mathcal{G}$ is formally defined as the Cartesian product of token probability distributions across all decoding steps $t=1$ to $T$:

$$
\mathcal{G} = \prod_{t=1}^T \Delta(\mathcal{V}) \subset \mathbb{R}^{|\mathcal{V}| \times T},
$$

where $\Delta(\mathcal{V})$ denotes the probability simplex over the vocabulary $\mathcal{V}$. This high-dimensional space encapsulates all possible output trajectories, with each element $\mathbf{g} \in \mathcal{G}$ representing a sequence of token probability distributions across decoding steps.

For LLMs, $\mathcal{G}$ is implicitly determined by the model architecture and is dynamically shaped by the employed decoding algorithms (e.g., greedy sampling, nucleus sampling). For sequences already generated by the LLM, we describe them as sequences obtained through sampling along the \textit{Generation Direction}.

\subsection{Cache Construction}

In CARD, we construct the target model’s cache by continuously inferring candidate tokens from the draft model. This approach leverages the demonstrated similarity between the draft model’s and target model’s output distributions, as validated by prior work such as EAGLE-2 \citep{Li2024EAGLE2FI} and OPT-tree \citep{wang2025opt}). To efficiently generate candidate tokens, the draft model employs a tree attention mechanism, selectively populating the cache with a subset of promising candidates. 

Managing the cache is challenging due to the extremely large vocabularies typical of modern LLMs. For instance, the LLaMA 3 series features a vocabulary size of 128,256 tokens. Storing all possible tokens in the cache is impractical for two main reasons: (1) it would impose an enormous memory overhead, and (2) even for the draft model, simultaneously processing all 128,256 tokens as input is computationally infeasible. Furthermore, excessively large input token sets can result in compute-bound scenarios that degrade the draft model’s inference speed. Empirical results indicate that caching a relatively small subset of tokens suffices to maintain a high mean acceptance length, enabling effective retrieval without excessive memory consumption. We provide a detailed analysis of cache size effects and its trade-offs in Section~\ref{ep:effectiveness_cache}.

We define the primary cache as $C^{(l)} = \{s_1^{(l)}, \dots, s_K^{(l)}\} \subseteq \mathcal{V}^l$. We also construct a \textbf{secondary cache, $\mathcal{B}^{(l)}$, to store the candidate tokens generated from the primary cache along with their corresponding path scores.} The primary cache size is $K$, where $K = |C^{(l)}|$, $l$ represents the current cache layer, and $\mathcal{V}$ denotes the vocabulary.

The cache construction process is illustrated in Figure 3. The draft model simultaneously processes all sequences from the current primary cache $C^{(l)}$ and computes the probability distributions over the next token in a single forward pass:
\begin{equation}
\{P_i^{(l)}\}_{i=1}^K = D_\phi(\text{TreeDecoding}(C^{(l)})) \tag{1}
\end{equation}
For each distribution $P_i^{(l)}$, we select the top-$k$ tokens to form the extension set:
\begin{equation}
T_i^{(l)} = \text{TopK}(P_i^{(l)}) = \{(y_{i,j}^{(l)}, p_{i,j}^{(l)})\}_{j=1}^k \tag{2}
\end{equation}
where $p_{i,j}^{(l)}$ represents the conditional probability of token $y_{i,j}^{(l)}$. To populate the secondary cache, the score of each new candidate is calculated by weighting its conditional probability with the cumulative probability of its parent sequence:
\begin{equation}
\forall(i, j), w_{i,j}^{(l)} = p_{i,j}^{(l)} \cdot \sigma(s_i^{(l)}) \tag{3}
\end{equation}
where $\sigma(s_i^{(l)})$ denotes the cumulative probability of the parent sequence $s_i^{(l)}$:
\begin{equation}
\sigma(s_i^{(l)}) = \prod_{t=1}^{l} p_\theta(s_i^{(l)}[t] \mid s_i^{(l)}[1:t-1]) \tag{4}
\end{equation}
The secondary cache must contain the candidate token, its full path score, and the index of its parent sequence to trace its origin. It is formally defined as a set of tuples:
\begin{equation}
\mathcal{B}^{(l)} = \{(y_{i,j}^{(l)}, w_{i,j}^{(l)}, i)\}_{i=1, j=1}^{K, k} \tag{5}
\end{equation}
After constructing the secondary cache, we reorder all $K \times k$ candidates globally based on their path scores. From this sorted pool, we select the top-$K$ candidates to form the new sequences for the next layer and update the primary cache. This ensures that only the most promising hypotheses are propagated. The process is defined as follows:
\begin{equation}
S^{(l)} = \text{Sort}(\mathcal{B}^{(l)}, \text{key=score, descending=True}) \tag{6}
\end{equation}
From the sorted list $S^{(l)}$, we select the top-$K$ candidates. Let the $m$-th best candidate be the tuple $(y_m^*, w_m^*, i_m^*)$, where $y_m^*$ is the token, $w_m^*$ is its full path score, and $i_m^*$ is the index of its parent sequence in $C^{(l)}$:
\begin{equation}
\{ (y_m^*, w_m^*, i_m^*) \}_{m=1}^K \subseteq S^{(l)} \tag{7}
\end{equation}
We then form the new sequences by appending each winning token to its corresponding parent sequence:
\begin{equation}
s_m^{(l+1)} = s_{i_m^*}^{(l)} \oplus y_m^* \quad \text{for } m=1, \dots, K \tag{8}
\end{equation}
Finally, the new primary cache for layer $l+1$ is formed by this set of $K$ new sequences:
\begin{equation}
C^{(l+1)} = \{s_1^{(l+1)}, \dots, s_K^{(l+1)}\} \tag{9}
\end{equation}
\begin{figure*}[t]
  \includegraphics[width=0.90\linewidth]{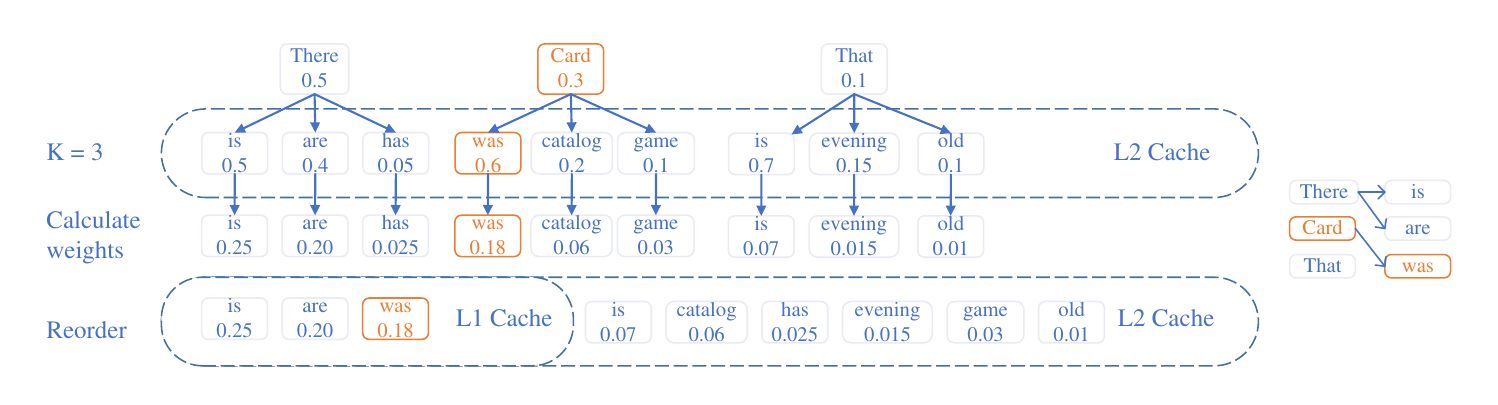}
  \caption{The process of constructing the cache.We use the first two layers of the cache construction outlined in our overview as an illustrative example.}
  \label{fig:Method_building_cache}
\end{figure*}

\subsection{Mask Construction for Draft Model}

CARD employs a draft model that generates a tree of candidate tokens in parallel. A key component enabling efficient parallel decoding across this tree is a dynamically constructed attention mask. The primary goal of this mask is to enforce the causal dependencies intrinsic to autoregressive generation while allowing multiple branches to be processed simultaneously.

\begin{figure}
  \includegraphics[width=\columnwidth]{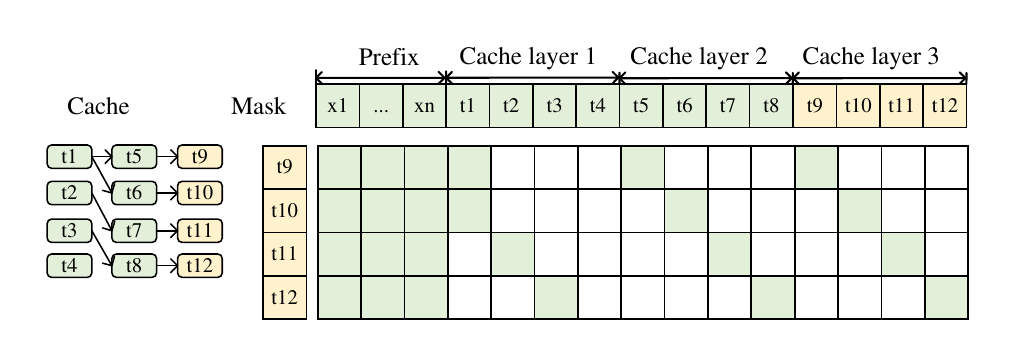}
  \caption{An example of mask construction: yellow nodes are the new cache to expand and the blue nodes are the existing cache.}
  \label{fig:method_mask}
\end{figure}

As illustrated in Figure~\ref{fig:method_mask}, at each decoding step $k$, the attention mask for the $N_k$ concurrently generated candidates is constructed according to two core principles:

\textbf{Causal History Inheritance.} Each candidate token must attend to all tokens along its ancestral path within the decoding tree. We realize this by programmatically inheriting the attention scope from each parent node, ensuring that each token's mask encompasses the full causal history accessible to its ancestor. This recursive mechanism guarantees that every node in the candidate tree maintains a valid causal context within the shared Key-Value (KV) cache.

\textbf{Intra-Step Independence.} For the $N_k$ tokens generated simultaneously at step $k$, we enforce mutual independence via an identity matrix within the attention mask. This design ensures that, conditioned on their shared history, each candidate is generated independently of its peers. Consequently, the generation of each candidate relies solely on its ancestral context, enabling their computations to proceed in parallel without information crosstalk.

The complete attention mask at step $k$ is thus formed by combining these two components: inheriting the full attention scope from the ancestors while constraining mutual attention among peers. This principled construction allows the draft model to explore multiple potential future trajectories concurrently, significantly accelerating inference while preserving the correctness of autoregressive decoding.

\subsection{Cache Correction}

The cache correction phase aims to realign the generation trajectory of the draft model with that of the target model. As depicted in Figure~\ref{fig:CARD_overview}, following verification by the target model, the validated tokens are propagated back to the draft model, which subsequently updates its cached information based on these tokens. This process involves pruning cache entries corresponding to incorrect decoding paths, thereby removing erroneous candidate sequences and freeing resources for more accurate candidates aligned with the correct generation direction.

Ideally, when paired with a suitably configured draft model, this correction mechanism enables the target model to perform inference at speeds approaching the draft model's throughput. The cache's retention of historical states allows the target model to retrieve and evaluate new candidate paths, even after rejecting previous proposed ones. For instance, as illustrated in Step 2 of Figure~\ref{fig:CARD_overview}, although the candidate token sequence "souvenir -" is rejected,  the new candidate path “of paper, it” remains accessible from the cache.

This framework attains comparable performance without requiring the target model to implement computationally expensive tree attention mechanisms for parallel verification of multiple sequences. In contrast to methods like OPT-Tree, Ouroboros, and EAGLE, which incur substantial computational overhead by incorporating tree attention in the target model, \textit{CARD effectively offloads the bulk of this computational burden to the draft model via its caching mechanism}. Consequently, CARD achieves comparable or superior acceleration while substantially reducing the computational demands placed on the target model.

\newcolumntype{d}{D{.}{.}{2}} 
\newcolumntype{C}{>{\centering\arraybackslash}X}

\begin{table*}
\centering
\small

\begin{tabular}{l l c c c c c c} 
\toprule
\multirow{2}{*}{Task} & \multirow{2}{*}{Framework} & 
\multicolumn{2}{c}{Llama3 70B/1B} & 
\multicolumn{2}{c}{Llama2 70B/7B} & 
\multicolumn{2}{c}{Llama2 70B/7B Temp=1.0} \\
\cmidrule(lr){3-4} \cmidrule(lr){5-6} \cmidrule(lr){7-8}
& & \multicolumn{1}{c}{tokens/s} & \multicolumn{1}{c}{speedup} 
  & \multicolumn{1}{c}{tokens/s} & \multicolumn{1}{c}{speedup} 
  & \multicolumn{1}{c}{tokens/s} & \multicolumn{1}{c}{speedup} \\
\midrule

\multirow{6}{*}{GSM8k} 
& Vanilla       & 8.75  & 1    & 8.80  & 1    & 8.80  & 1    \\
& Speculative   & 21.22 & 2.42 & 16.68 & 1.89 & 16.57 & 1.88 \\
& Lookahead     & -     & -    & 13.52 & 1.53 & 13.65 & 1.55 \\
& Ouroboros     & -     & -    & 23.45 & 2.66 & 19.27 & 2.19 \\
& PEARL         & 25.76 & 2.94 & 20.34 & 2.27 & 18.74 & 2.12 \\
& \textbf{CARD} & \textbf{41.34} & \textbf{4.72} & \textbf{29.81} & \textbf{3.39} & \textbf{20.30} & \textbf{2.30} \\
\midrule

\multirow{6}{*}{MGSM} 
& Vanilla       & 9.48  & 1    & 9.64  & 1    & 9.64  & 1    \\
& Speculative   & 20.32 & 2.14 & 17.21 & 1.78 & 17.14 & 1.77 \\
& Lookahead     & -     & -    & 14.31 & 1.48 & 14.45 & 1.49 \\
& Ouroboros     & -     & -    & 25.02 & 2.59 & 23.97 & 2.48 \\
& PEARL        & 28.68 & 3.02 & 22.88 & 2.37 & 21.00 & 2.17 \\
& \textbf{CARD} & \textbf{37.93} & \textbf{4.00} & \textbf{30.56} & \textbf{3.17} & \textbf{26.50} & \textbf{2.74} \\
\midrule

\multirow{6}{*}{MT-Bench} 
& Vanilla       & 9.18  & 1    & 9.38  & 1    & 9.38  & 1    \\
& Speculative   & 17.73 & 1.93 & 14.89 & 1.59 & 16.78 & 1.79 \\
& Lookahead     & -     & -    & 14.02 & 1.49 & 14.01 & 1.49 \\
& Ouroboros     & -     & -    & 23.68 & 2.52 & 21.84 & 2.32 \\
& PEARL        & 22.97 & 2.50 & 20.27 & 2.16 & 18.99 & 2.02 \\
& \textbf{CARD} & \textbf{28.32} & \textbf{3.08} & \textbf{25.97} & \textbf{2.76} & \textbf{25.20} & \textbf{2.68} \\
\bottomrule
\end{tabular}
\caption{Performance comparison across different tasks and architectures}
\label{tab:performance}
\end{table*}

\section{Evaluation}

This section presents a systematic evaluation of CARD on diverse text generation benchmarks, focusing on computational efficiency and performance stability across different architectural configurations.

\subsection{Experimental Setup}

\textbf{Models and Datasets}. 
For code generation, we evaluate CARD on the HumanEval \citep{chen2021evaluating} and MBPP \citep{austin2021program} datasets. HumanEval comprises 164 entries, each consisting of a text prompt and a prefix of a Python function. The MBPP test set contains 500 entries, requiring the model to generate entire functions based on prompts and test cases. From these, we randomly sampled 100 entries for evaluation.
For arithmetic reasoning, multi-round conversation, and multilingual tasks, we utilize GSM8K \citep{cobbe2021training}, MT-Bench \citep{zheng2023judging}, and MGSM \citep{shi2022language} datasets. For these evaluations, 100 entries are randomly sampled from GSM8K. Across all datasets, the maximum number of tokens generated per instance was set to 512.For HumanEval and MBPP, we experiment with three draft-target model pairs:
\begin{itemize}
\item \textbf{Group 1:} LLaMA-2-chat-7B (draft) and LLaMA-2-chat-70B (target) \citep{touvron2023llama}.
\item \textbf{Group 2:} Qwen-2.5-7B (draft) and Qwen-2.5-70B (target) \citep{qwen2.5}.
\item \textbf{Group 3:} LLaMA 3.2-1B (draft) and LLaMA 3.2-70B (target) \citep{grattafiori2024llama}.
\end{itemize}
For GSM8K, MT-Bench, and MGSM, the first and third model groups are employed to assess the impact of different draft models within the CARD framework. This setup enables us to evaluate how the choice of draft model influences inference efficiency and overall performance.

\textbf{Baselines.} CARD is a general approach compatible with any SD framework. For comparison, we select several training-free baseline methods, including vanilla autoregressive decoding, speculative decoding, lookahead decoding \citep{fu2024break}, Ouroboros  \citep{}, and PEARL  \citep{}, all evaluated under their default configurations. We exclude specially trained drafters such as EAGLE-3 \citep{li2025eagle}, which achieve superior drafting performance due to extensive training, making direct comparisons with such specialized models unfair.

Following our \textit{“query-and-correct”} paradigm, we define the \textit{communication ratio} as the ceiling of the ratio between the forward computation times of the target and draft models. This ratio controls how frequently the target model queries the cache and corrects the draft model’s generation direction, with the goal of maximizing draft model throughput while minimizing idle time. In all experiments, for 7B models, the communication ratio is set to 5:1, meaning the draft model generates five token sequences for every query and correction by the target model. We set the cache size parameter $K=50$. For 1B draft models, the communication ratio is increased to 7:1 with $K=100$. Unless otherwise specified, the temperature parameter is fixed at 0 for all decoding methods.

\textbf{Metrics.} Our evaluation focuses on three metrics: (1) \textit{Speedup Ratio}, measuring acceleration over standard autoregressive decoding; (2) \textit{Mean acceptance length}, the average number of tokens accepted per decoding step, limited by the drafter’s prediction accuracy and execution speed in CARD; (3) \textit{Computational Efficiency}, a measure designed to capture the computational cost of verification. We define this efficiency metric as the product \(\text{Params} \times L_{\text{new}}\), where \(\text{Params}\) is the model size in billions, indicating complexity and decoding cost, and \(L_{\text{new}}\) is the number of tokens processed per forward pass. This product effectively quantifies the computational efficiency during single-step verification.

\textbf{Hardware and Implementation.} Experiments are conducted on three NVIDIA A800 80GB GPUs connected via NVLink×8 and an Intel Platinum 8350C CPU. We utilize HuggingFace Transformers v4.31.0 for all procedures to ensure compatibility and reproducibility.
\begin{figure*}[t]
    \includegraphics[width=0.45\linewidth]{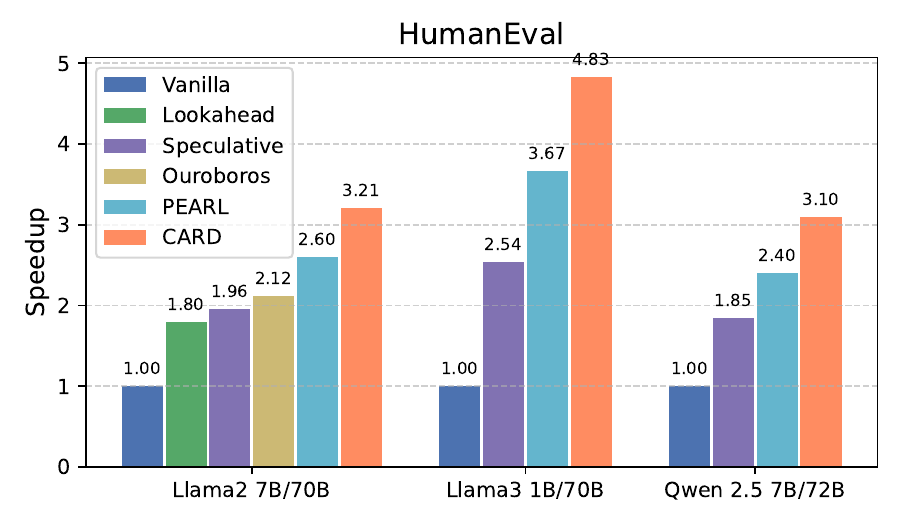} \hfill
    \includegraphics[width=0.45\linewidth]{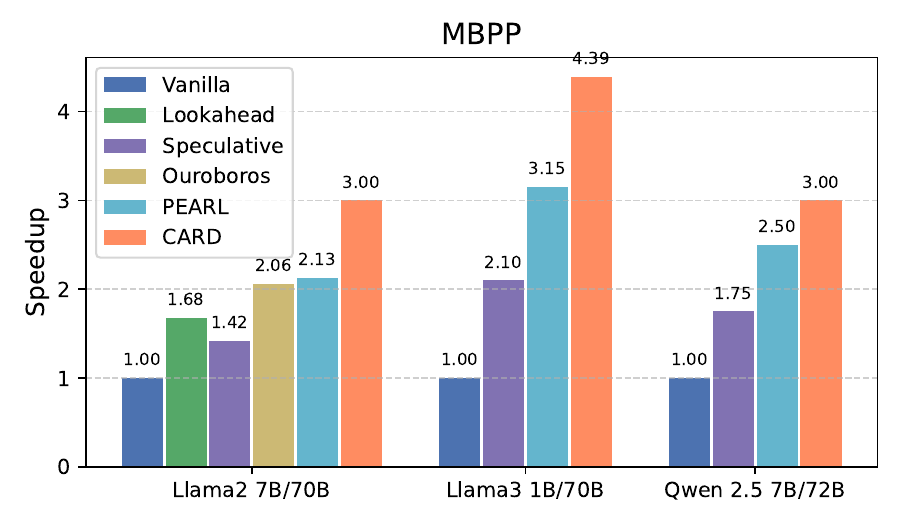}
    \caption{The greedy decoding speedup on HumanEval and MBPP}
    \label{fig:ep_multi_model}
\end{figure*}
\subsection{Overall Results}

\paragraph{Multi-model Performance.} As shown in Figure~\ref{fig:ep_multi_model}, on HumanEval and MBPP, CARD consistently surpasses other baselines across models of different sizes, achieving a maximum speedup of 4.83× on the Llama 3 1B/70B models. However, when employing the 7B model as the draft model, the acceleration ratios are less pronounced, primarily constrained by the inference speed of the draft model itself. Detailed inference speeds for models of various sizes are provided in the Appendix. On GSM8K, MT-Bench, and MGSM, Table~\ref{tab:performance} shows that CARD achieves the highest speedup ratios across all tested settings, demonstrating robust advantages over existing training-free SD frameworks.

\paragraph{State-of-the-art Acceleration Performance.} CARD outperforms the vanilla autoregressive decoding baseline by 4.72× on GSM8k, 4.00× on MGSM, and 3.08× on MT-Bench under the Llama 3 architecture. It consistently maintains leading speedup ratios under Llama 2 configurations (3.39×/3.17× for GSM8k/MGSM).

\paragraph{Temperature-Robust Speedup.} Across the three datasets presented in Table~\ref{tab:performance}, CARD exhibits some performance decline when the temperature is set to 1. However, it still outperforms other baselines, showcasing its ability to effectively balance generation quality and acceleration under diverse sampling conditions.

\paragraph{Superior Performance with Smaller Draft Models.} Our experiments reveal that the choice of draft model plays a crucial role in CARD’s acceleration performance. When employing Llama3-1B as the draft model, CARD achieves a peak speedup of 4.72×, significantly outperforming the 3.39× speedup observed with the larger Llama2-7B model. This counterintuitive outcome demonstrates the effectiveness of our \textit{``query-and-correct''} architecture.

The key trade-off lies between the draft's quality and the overall generation speed. Although larger models, such as the 7B model, produce higher-quality candidates, their inherently slower inference creates a bottleneck. In contrast, the smaller 1B model, despite generating lower-quality candidates, enables faster drafting. Thanks to CARD's \textit{``query-and-correct''} mechanism, complemented by its cache component, it effectively compensates for the draft's lower quality, allowing for high acceptance lengths without incurring the latency typical of larger models. Notably, the Llama3-1B model used here is not specifically optimized for this task. Employing a dedicated draft model trained for this purpose is anticipated to further boost performance.

\subsection{Ablation Study}
\begin{table}[t]
\centering
\begin{tabular}{l r}
\toprule
Method & tokens/s \\
\midrule
Baseline & 8.75 \\
+ Cache (query but don't correct) & 30.23 \\
+ Correct (correct generation direction) & 41.34 \\
\bottomrule
\end{tabular}%
\caption{The ablation studies of each component in CARD on GSM8k using Llama3 1B/70B}
\label{tab:ep_ablation}
\end{table}
As shown in Table~\ref{tab:ep_ablation}, we perform ablation experiments on the GSM8k dataset using both Llama3-1B and 70B models to evaluate the contributions of our framework components. When only the cache component is incorporated, the target model can perform queries but does not correct the draft model. This setup reveals an inherent instability in the draft model's directional prediction capability, as longer generation sequences tend to amplify discrepancies between the predicted and verified token distributions. Nonetheless, the inherent parallelism and the abundance of candidate tokens provided by the cache still enable reasonable speedup ratios. Upon adding the correction component, the issue of divergence in generation direction predictions is significantly alleviated, resulting in notable performance improvements.

\subsection{The Effectiveness of Cache Size (K)} 
\label{ep:effectiveness_cache}
\begin{figure*}
    \centering
    \includegraphics[width=0.32\linewidth]{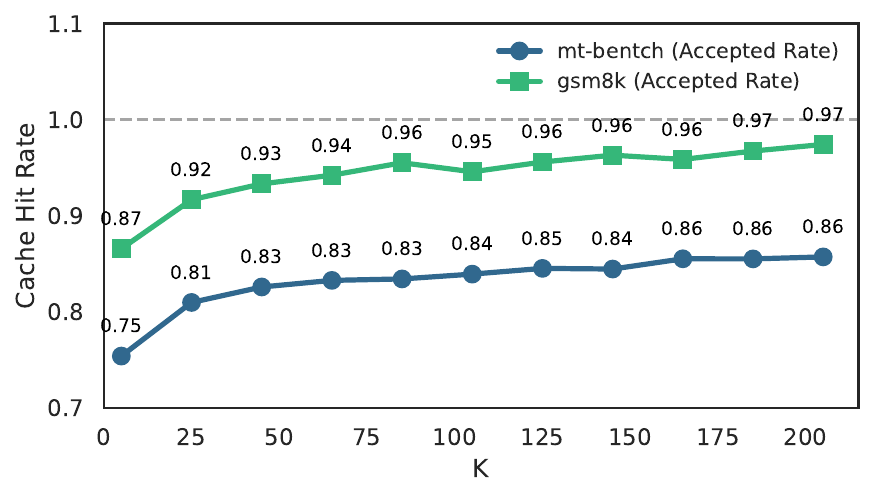} \hfill
    \includegraphics[width=0.32\linewidth]{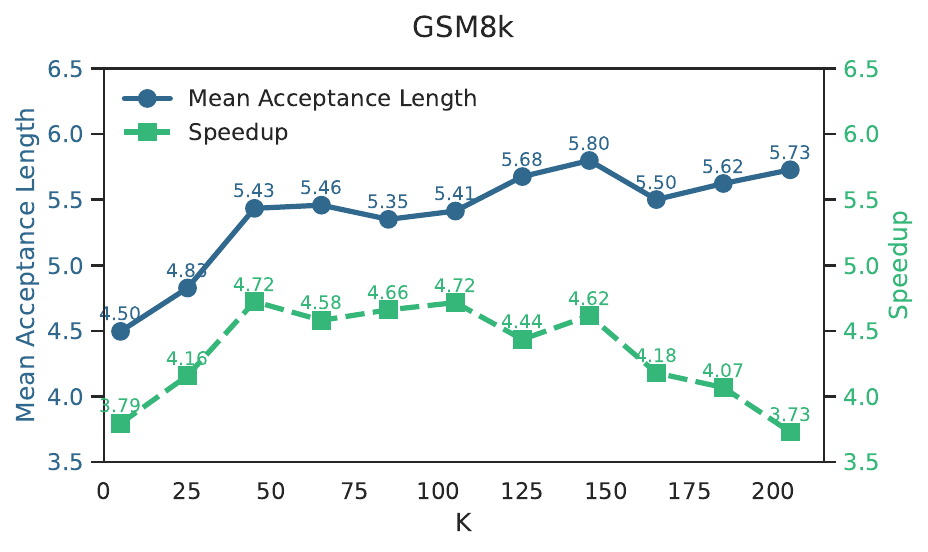} \hfill
    \includegraphics[width=0.32\linewidth]{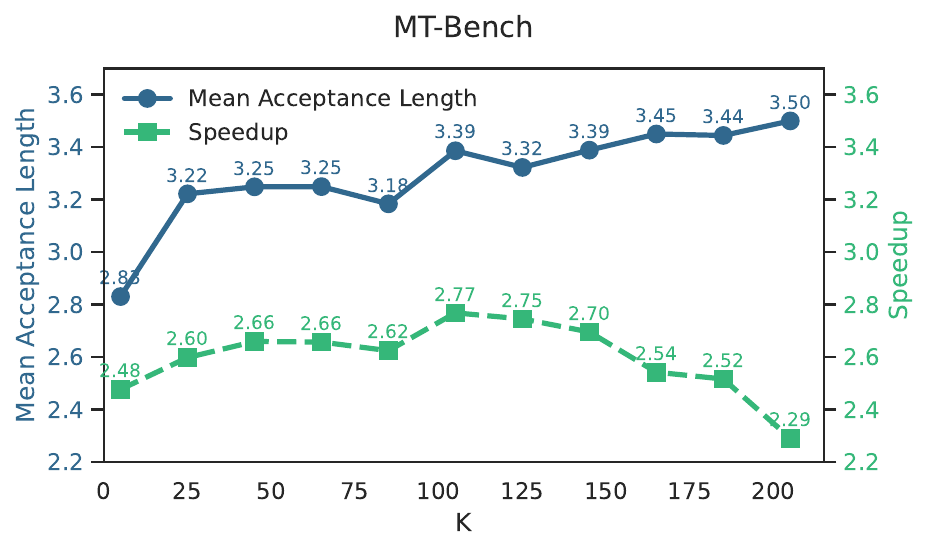}
    \caption{Left: Cache hit rate variation with K; Middle: Performance on GSM8K (mean acceptance length and tokens per second) with K; Right: Performance on MT Bench (mean acceptance length and tokens per second) with K}
    \label{fig:ep_effectiveness_cache}
\end{figure*}

As shown in Figure~\ref{fig:ep_effectiveness_cache}, we examine the impact of varying the cache size parameter K on performance across the MT-Bench and GSM8k datasets. With the communication ratio fixed at 7, we incrementally increase K from 5 in steps of 25, and tracking the mean acceptance length, tokens per second, and cache hit rate. The results indicate that on both datasets, the mean acceptance length increases with larger K values, suggesting that bigger caches offer greater benefits to the target model. However, beyond K=50, the improvements become gradually less pronounced. Interestingly, the speedup exhibits an initial increase followed by a decline, implying that once the cache size surpasses a certain threshold, the draft model becomes compute-bound, which subsequently limits overall speed. The cache hit rate trends upward with increasing K, reaching an impressive maximum of 97\%. This high hit rate confirms the effectiveness of our correction process, which ensures the draft model consistently predicts along the target model’s generation direction,  enabling  exceptional cache utilization.

\subsection{Comparing with Tree-Verified Methods}

CARD achieves performance comparable to traditional tree-based verification methods while significantly reducing the primary model's computational load. This efficiency stems from its \textit{“query-and-correct”} mechanism, which delegates most token verification tasks from the resource-intensive target model to a lightweight draft model. As a result, the target model's computational resources are freed up for more efficient inference.
Quantitative evaluations on the GSM8k dataset verify this advantage. 

As shown in Table~\ref{tab:ep_efficiency}, using the Params×Lnew metric (which combines model size and the number of new tokens processed), CARD matches the performance of advanced methods such as EAGLE-3 while substantially reducing the computational burden on the target model. This efficiency gain is largely due to CARD's communication design: with a communication ratio of 7, the target model verifies approximately 7 candidate tokens per step (though this number may fluctuate slightly due to cache misses). Importantly, by effectively offloading workload to the draft model, CARD enables processing more input tokens with fewer target model resources. This optimized resource allocation makes CARD especially valuable in resource-constrained environments.

\begin{table}[t]
\centering
\small

\begin{tabular}{@{}lrrr@{}} 
\toprule
Method & \multicolumn{1}{c}{Target Model} & \multicolumn{1}{c}{Draft Model} & Speedup \\
       & \multicolumn{1}{c}{Computation} & \multicolumn{1}{c}{Computation} & \\
\midrule
OPT-Tree & 50$\times$70 (3500) & 7$\times$50 (350) & 1.9 \\
EAGLE-3 & 48$\times$70 (3360) & \textbf{1$\times$10 (10)} & 4.43 \\
CARD & \textbf{7$\times$70 (490)} & 1$\times$100 (100) & \textbf{4.72} \\
\bottomrule
\end{tabular}
\caption{Comparison of computational efficiency on GSM8k with 70B target models.EAGLE-3 and OPT-Tree use fixed tree sizes: 50 for OPT-Tree and 48 for EAGLE-3. In the draft model comparison, OPT-Tree uses a 7B model, while EAGLE-3 and CARD use 1B models.}
\label{tab:ep_efficiency}
\end{table}

\section{Related Work}

LLM inference acceleration  has been actively explored through a variety of complementary approaches. Decoding algorithm research includes non-autoregressive decoding frameworks \citep{guo2020jointly,ghazvininejad2019mask} that generate multiple tokens in parallel, and draft-then-verify methods \citep{stern2018blockwise,xia2022speculative} that leverage speculative sampling to achieve lossless acceleration. Variants of speculative decoding exploit small draft models \citep{leviathan2023fast, zhou2023distillspec}, internal target-model mechanisms \citep{cai2024medusa,fu2024break}, and external phrase retrieval techniques \citep{saxena2023prompt,he2023rest}. Additionally, tree-style verification approaches enable parallel validation of draft token sequences \citep{miao2024specinfer,cai2024medusa}. On the implementation side, hardware-aware optimizations target memory-efficient attention computation \citep{dao2023flashdecoding} and distributed model execution \citep{shoeybi2019megatron}, improving throughput on modern architectures. Model compression techniques complement these methods by reducing computational demands through quantization \citep{frantar2022gptq}, pruning \citep{han2015learning}, and distillation \citep{hinton2015distilling}. These orthogonal directions, such as  decoding strategies, system-level optimizations, and model compression, are often combined to maximize inference acceleration in practice.

\section{Conclusion}

In this paper, we propose CARD, featuring a novel paradigm called \textit{``query-and-correct''} for speculative decoding. Our approach addresses the intrinsic mutual waiting bottleneck present in traditional ``draft-then-verify'' schemes by employing a cache-assisted parallel speculative decoding strategy. By integrating a cache as middleware between the draft model and the target model, we enable the draft model to effectively predict the target model's generation space. This prediction allows us to offload substantial computational burden from the target model onto the draft model, while preserving high mean acceptance lengths during decoding. Extensive evaluations demonstrate that CARD significantly outperforms state-of-the-art SD baselines without requiring any additional training or fine-tuning, highlighting its superiority for efficient LLM serving.

\bibliography{aaai2026}


\end{document}